\pdfoutput=1

\documentclass[11pt]{article}

\usepackage[]{naacl2021}

\usepackage{times}
\usepackage{latexsym}
\usepackage{tabu,multirow}

\usepackage{comment}

\usepackage[T1]{fontenc}

\usepackage[utf8]{inputenc}
\usepackage{graphicx}
\usepackage{amsmath}

\usepackage{microtype}

\definecolor{sm}{HTML}{FF0000}

\title{ENTRUST: Argument Reframing with \\Language Models and Entailment}

\author{Tuhin Chakrabarty\textsuperscript{1}, 
  \textbf{Christopher Hidey}   \textsuperscript{3}
 \textbf{Smaranda Muresan}\textsuperscript{1,2}\\ 
  Department of Computer Science, Columbia University \textsuperscript{1} \\
    Data Science Institute, Columbia University \textsuperscript{2}\\
    Google \textsuperscript{3} \thanks{~~The work is not affiliated to Google and was conducted independently outside of the organization}\\
  {\tt \{tuhin.chakr, smara\}@cs.columbia.edu}\\
  {\tt chris.hidey@gmail.com}
  }

\begin{document}
\maketitle
\begin{abstract}

Framing involves the positive or negative presentation of an argument or issue depending on the audience and goal of the speaker \cite{Entman1983FramingTC}. Differences in lexical framing, the focus of our work, can have large effects on peoples’ opinions and beliefs. To make progress towards reframing arguments for positive effects, we create a dataset and
method for this task. We use a lexical resource for connotations to create a parallel corpus and propose a method for argument reframing that combines controllable text
generation (positive connotation) with a postdecoding entailment component (same denotation). Our results show that our method is effective compared to strong baselines along the
dimensions of fluency, meaning, and trustworthiness/reduction of fear.

\end{abstract}

\section{Introduction}
Public opinion has been shown to be significantly influenced by \textit{framing effects.} Framing refers to the presentation of an issue, where even small changes may have outsized effects on beliefs \cite{Chong2007FramingT}. For example, when asked about ``welfare,'' the American public is largely against increasing spending (with only 20\% in favor), but when asked about ``assistance to the poor,'' 65\% believe that the government is not spending enough \cite{Rasinski1989THEEO}.

 While other research has focused on \textit{syntactic} framing \cite{greene-resnik-2009-words} or \textit{issue} framing \cite{hartmann-2019-predicting}, we focus specifically on \textit{lexical} framing, distinguishing sentences by their \textit{connotative} meaning even where they have the same \textit{denotative} meaning. According to \newcite{FregeberSU}, two sentences with the same truth conditions may refer to the same entities or state of affairs (\textit{``reference,''} also known as denotation) but be presented differently (\textit{``sense''} or connotation).
For example, ``undocumented workers'' and ``illegal aliens'' have the same denotation but different connotations \cite{webson-etal-2020-undocumented}.

\begin{table}[t]
\small
\centering
\begin{tabular}{|l|l|}
\hline
Arg1                                              & \begin{tabular}[c]{@{}l@{}}Alabama's Supreme Court Chief Justice was \\suspended... for ordering state probate\\ judges not to grant \textit{\color{red}marriage licenses} to gay  \\couples...\end{tabular}    \\ \hline
\begin{tabular}[c]{@{}l@{}}Rf\\ Arg1\end{tabular} & \begin{tabular}[c]{@{}l@{}}Alabama's Supreme Court Chief Justice was \\suspended... for ordering state probate\\ judges not to grant \textit{\color{blue}legal marriage  equality} \\to gay couples...
\end{tabular} \\ \hline\hline
Arg2                                              & \begin{tabular}[c]{@{}l@{}}Every nation with \textit{\color{red}territorial claims} in the\\  arctic is a member of NATO, except Russia.\end{tabular}          \\ \hline
\begin{tabular}[c]{@{}l@{}}Rf\\ Arg2\end{tabular} & \begin{tabular}[c]{@{}l@{}}Every nation with \textit{\color{blue}sovereign competence} in the\\ arctic is a member of NATO, except Russia.\end{tabular}         \\ \hline\hline 
Arg3                                              & \begin{tabular}[c]{@{}l@{}}At this \textit{\color{red}dire} moment , we all need to amplify our \\voices in \textit{\color{red}defense} of free speech .
\end{tabular}                              \\ \hline
\begin{tabular}[c]{@{}l@{}}Rf\\ Arg3\end{tabular} &  \begin{tabular}[c]{@{}l@{}}At this \textit{\color{blue}crucial} moment , we all need to amplify\\ our voices in \textit{\color{blue}support} of free speech.
\end{tabular}                              \\ \hline\hline
Arg4                                              & \begin{tabular}[c]{@{}l@{}}It is difficult to think of any single act that\\ would do more to restore America's \textit{\color{red}soft power}\\ than the election of Obama to the presidency\end{tabular}                              \\ \hline
\begin{tabular}[c]{@{}l@{}}Rf\\ Arg4\end{tabular} &  \begin{tabular}[c]{@{}l@{}}It is difficult to think of any single act that\\ would do more to restore America's \textit{\color{blue}diplomatic}\\ \textit{\color{blue}credibility} than the election of Obama to the \\presidency\end{tabular}                              \\ \hline
\end{tabular}
\caption{Examples of arguments (Arg1, Arg2) with  high partisan skew collocations (in red) \cite{webson-etal-2020-undocumented} as well as appeal to fear or prejudice argument fallacies (Arg3, Arg4) \cite{da2019fine}, along with reframed arguments as an attempt by our model \textbf{ENTRUST}  to improve trustworthiness.}
\label{table:example1}
\end{table}

The examples in Table \ref{table:example1} are instances of \textit{lexical} framing, where word choice determines the difference in presentation \cite{Mccombs2001TheCO}. For example, Arg1 and Arg2 contain collocations (in red) that have a high partisan skew \cite{webson-etal-2020-undocumented}, while Arg 3 and Arg 4 are examples of appeal to fear or prejudice argument fallacies from propagandist news articles \cite{da2019fine}. The goal is to reframe such arguments to be more trustworthy (e.g., less partisan, no appeal to fear fallacy).

Connotations may be distinguished along the dimensions of politeness, sentiment, or tangibility, among others \cite{allaway2020unified}, but in our work we consider emotional association such as fear and trust. Appeal to \textit{fear} is considered an argumentative fallacy \cite{Walton2006FundamentalsOC, Thierer2012TechnopanicsTI} and appears  prominently in manipulative text such as propaganda \cite{da2019fine}.
On the other hand, arguments with \textit{trusted} language align with the Aristotelian modes of persuasion, specifically ethos \cite{Aristotle2019AristotlesO}.

In our work, we leverage such a lexical resource for connotations \cite{allaway2020unified} to reframe arguments to be more trustworthy (e.g., less partisan, no appeal to fear fallacy), while maintaining the same denotative meaning. While retrieve-and-replace methods perform well on other attribute transfer tasks such as sentiment \cite{li2018delete,sudhakar2019transforming}, our task is more dependent on broader context within a sentence even though we are performing localized replacement. Thus, there are two main challenges we need to address: 1) the lack of a parallel dataset of negatively and positively framed arguments (naturally-occurring); and 2) a generation approach that can not only change the connotative meaning but also keep the same denotative meaning of the input argument.

We introduce our approach called \textbf{ENTRUST }: Argum\textbf{ENT} \textbf{R}eframing with lang\textbf{U}age model\textbf{S} and en\textbf{T}ailment, with the following contributions: 1) A Connotation-guided Masked Language Model approach to generate a parallel dataset of naturally occuring arguments and their reframings (Section \ref{section:data1}); 2) A method for argument reframing that combines controllable text generation (connotative meaning associated with trust) and entailment (same denotative meaning) (Section \ref{sec:method});   
3) An evaluation on two different tasks --- reframing partisan arguments and appeal to fear/prejudice fallacies --- showing that our method is preferred over a strong retrieval-based baseline \cite{sudhakar2019transforming} and state-of-the-art pretrained language model \cite{lewis2019bart}, and it is close to human performance on several evaluation criteria such as fluency, meaning, trustworthiness/reduction in fear. Code, data, and models available at \url{https://github.com/tuhinjubcse/ArgReframingNAACL2021}  

\section{Automatic Parallel Data Creation} \label{section:data1}
To facilitate the reframing of arguments, we require a large-scale parallel corpora of sentences with the same denotation but different connotative meaning.

\paragraph{Selection of naturally-occurring arguments.} Since our goal is to re-write arguments, it is essential to identify an abundant source of naturally-occurring arguments. The \textit{Change My View} subreddit, an argumentative discussion forum intended for persuasion on diverse topics, has been used extensively in computational argumentation research \cite{tan2016winning,wei2016post,musi2018changemyview,chakrabarty2019imho, chakrabarty2019ampersand,hidey2017analyzing}. We collect sentences from the same source and classify them as \textit{claim}, \textit{premise}, or \textit{non-argument} using the fine-tuned BERT model released by \citet{chakrabarty2019ampersand}. This results in 301,166 arguments labeled as premises. We consider only premises to create our parallel data because argumentative appeals occur within justifications (premises) for or against the speaker's claim.

\paragraph{Connotation-guided Masked Language Model.} \citet{allaway2020unified} provide a resource with words labeled for lexical connotations, using the aspects of \textit{Social
Value, Politeness, Impact, Factuality, Sentiment, Emotional Association}. For our work we only consider \textit{Emotional Association}, although in future work our methods could be applied for other aspects. 
To create a parallel corpus, we use this lexical resource and the 301,166 automatically identified premises from \textit{Change My View} to obtain candidate words within those premises for replacement.  We match words from the premises to those that have  entries in the dictionary with \textit{emotional} connotations such as fear, trust, anticipation, and joy. To generate replacements for these words, we need to find substitutions that maintain denotative meaning while changing connotative meaning. We use the connotation dictionary to address the latter. However, to address the former, we need to provide only paraphrases that consider the context in which these words occur. We thus use a masked language model (MLM).

\begin{figure*}
\centering
\includegraphics[scale=0.65]{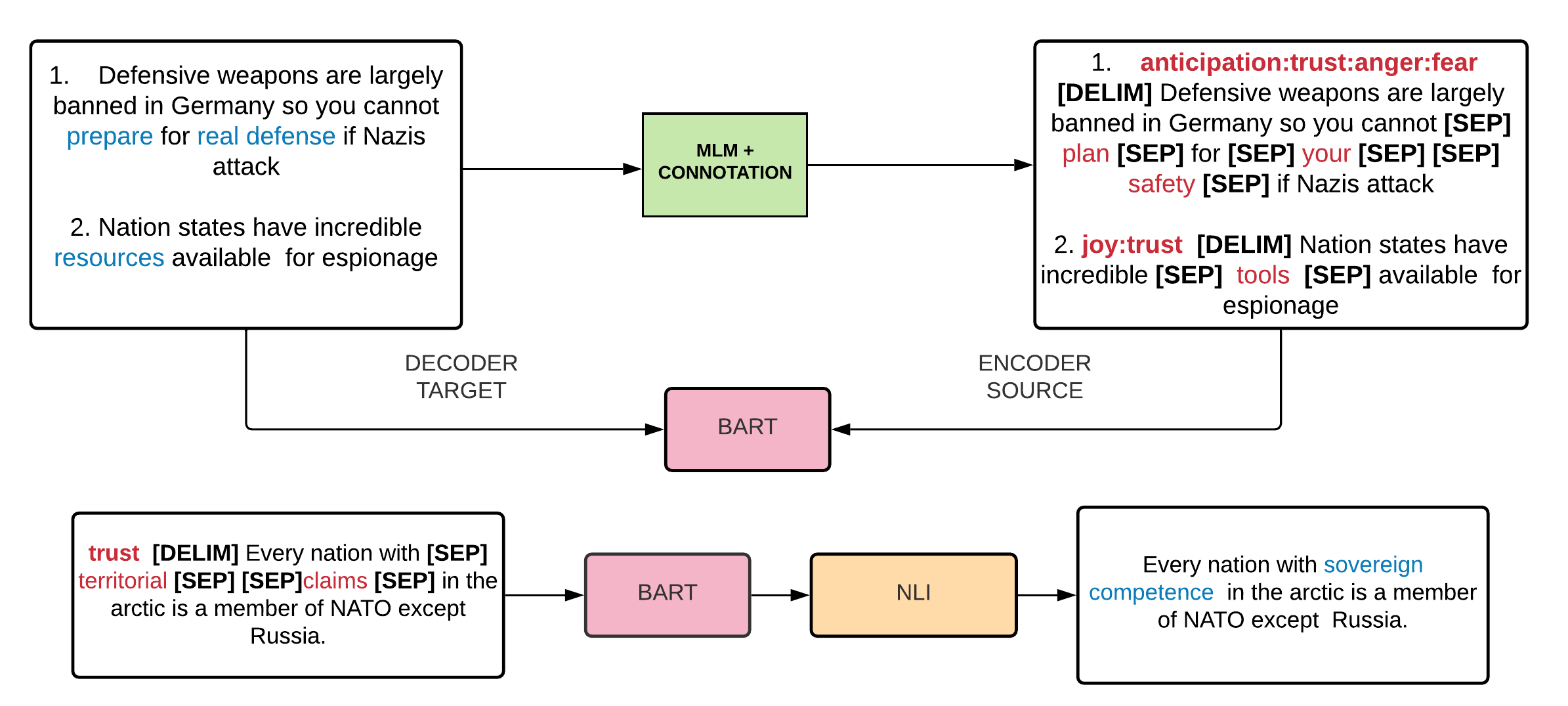}
\caption{\label{figure:sim} A schematic illustration of our system ENTRUST, where the top block shows our \textbf{training} process where we use MLM along with connotation resource to transform an original argument to an argument bearing different emotional connotation and use them to fine-tune BART. The block below shows the \textbf{inference} step (test time) where we use fine-tuned BART to reframe the argument containing \textit{partisan collocation} along with a NLI component to ensure the same denotation with the input argument.}
\vspace{-.5em}
\end{figure*}

Masked language modeling approaches like BERT \cite{devlin2018bert} and RoBERTa \cite{liu2019roberta} can be considered \textit{cloze} or ``fill-in-the-blank'' tasks, where the model uses the context  surrounding a masked-out token to try to predict what the masked word should be. We borrow this framework (RoBERTa-large, in particular) to mask the candidate words we identified via the connotation lexicon. However, the rank of a predicted token from an MLM is based on the  language model probability -- it provides no information about lexical connotations. A premise re-written from MLM replacements may thus have the same connotative meaning. To avoid this scenario, we restrict the MLM replacements to be words with different connotations than the original masked word (i.e., different Emotional Association).

Our data creation process is depicted in Figure \ref{figure:sim}. In example 2, the word \textit{``resources''} has the connotations \textit{joy;trust} in our dictionary. The MLM generates the replacement \textit{``tools,''} which we verify has a different connotation (\textit{emotionally neutral}).
For example 1, the words \textit{"prepare,"} \textit{"real,"} and \textit{"defense"}  have the emotional connotations \textit{anticipation}, \textit{trust}, and \textit{anticipation;anger;fear} respectively. These words are replaced with \textit{plan}, \textit{your}, and \textit{safety}, using our MLM.

We treat the original premises as the ``target'' and  the connotation-guided MLM generated premises  as the ``source" for our method of argument reframing detailed in the next section (Figure \ref{figure:sim}). 
 While this process provides us with a parallel dataset for reframing, we enhance the source-side of the data to provide additional control during generation.
Motivated by the work of \citet{schiller2020aspect}, which used aspect as a ``control code''  \cite{keskar2019ctrl} for argument generation, we also prepend the emotional associations of the replaced words. Using the connotations from the lexical resource, we add all listed emotions as control codes by separating them with a special token (``[DELIM]'') (the top right block of Figure \ref{figure:sim}). During inference, we thus have more control over the emotion of the words we are generating (in our case we specifically use \text{trust} as the control code). For additional control, we also insert demarcator tokens (``[SEP]'') at the boundary of the words we aim to replace to provide our generative model with a better signal on what to replace or rewrite. While the downside is that we need to identify spans for replacement at test/inference time, our experiments will show that using collocations or fear words makes it unnecessary.

By using the lexical connotation resource 
we do not have to rely on a separate module/tagger based approach like that of \citet{pryzant2020automatically} to find biased or problematic words that may introduce additional noise during training. Our parallel data has 271,022 pairs for training and 30,114 for validation on which  perplexity is evaluated.

\section{Method for Argument Reframing} \label{sec:method}
As our goal is to change \textit{connotation} while maintaining \textit{denotation}, we divide our approach to re-writing arguments into two primary tasks: 1) generating the appropriate lexical substitutions while being pertinent to the context; 2) ensuring that re-written arguments reflect the desired emotional association while maintaining the same denotative meaning as the input.

\subsection{Controllable Text Generation}
BART \cite{lewis2019bart} is a pre-trained model combining bidirectional and auto-regressive transformers that achieves state-of-the-art results in several text generation tasks. It is implemented as a sequence-to-sequence model with a bidirectional encoder over corrupted text and a left-to-right auto-regressive decoder. In principle, the pre-training procedure has two stages: (1) text is corrupted with an arbitrary noising function, and (2) a transformer-to-transformer model is learned to reconstruct the original text. Because BART has an auto-regressive decoder, it can be directly fine-tuned for most sequence generation tasks. Here, the encoder input is a sequence of words, and the decoder generates outputs auto-regressively. We refer the reader to \cite{lewis2019bart} for further details.  

For our task, we fine-tune BART on our parallel data, where the reframed argument using MLM \& connotation dictionary is the encoder source and the original argument is the decoder target (Figure \ref{figure:sim}). The emotional connotations added to the source via  the special token DELIM (see Section \ref{section:data1}) act as a ``control code'' for generation.  Moreover, for lexical framing, subtle differences in word choices matter the most.
By explicitly using special tokens ([SEP]) in our parallel data during fine-tuning, the BART model learns what to edit, instead of editing random words in the sentence, a common issue often found in attribute transfer models \cite{li2018delete,sudhakar2019transforming}. At test time, therefore, we can ensure the model reframes a desired content span. All hyper-parameters are mentioned in the \textbf{Appendix A}.

Post fine-tuning at the decoding step, we use a top-k sampling strategy \cite{fan2018hierarchical} to reframe arguments conditioned on a input argument and a target emotion.

\begin{table}[]
\small
\centering
\begin{tabular}{|l|l|}
\hline
Source                                               & \begin{tabular}[c]{@{}l@{}}trust \textless{}V\textgreater I suppose we could argue that they're \\much better at \textit{\color{red}soft power} than Nazi Germany \\or the USSR, but come on\end{tabular} \\ \hline
BART                                                 & \begin{tabular}[c]{@{}l@{}}I suppose we could argue that they're much \\better at \textit{\color{blue}military strength} than .......... \end{tabular}                               \\ \hline
\begin{tabular}[c]{@{}l@{}}BART\\ + NLI\end{tabular} & \begin{tabular}[c]{@{}l@{}}I suppose we could argue that they're much\\ better at \textit{\color{teal}diplomatic communication} than ......\end{tabular}                                 \\ \hline
\end{tabular}
\caption{Generation from fine-tuned BART without control for entailment can sometime contradict the input thereby failing to maintain the same denotative meaning}
\label{table:contradict}
\vspace{-4ex}
\end{table}

\subsection{Post-decoding NLI}
 Our task is 
 challenging in comparison to traditional text attribute transfer tasks as we need to maintain the same denotative meaning as the input. While in most cases BART is able to generate content which is semantically similar to the input, it sometimes contradicts the input. For example, Table \ref{table:contradict} shows that BART changes \textit{soft power} to \textit{military strength}. Here the denotative meaning changes. To control for this, we introduce an additional post-processing step.

We generate multiple outputs by varying the value of \textit{k} (between 5 and 50) while conducting top-k sampling. We then calculate the entailment scores of these outputs with the input argument respectively using a RoBERTa \cite{liu2019roberta} model fine-tuned on the Multi-NLI dataset \cite{N18-1101} and then select the output having the best entailment score. We also experimented with other methods for incorporating entailment during decoding based on prior work (Section \ref{sec:related}), but found these techniques to be less effective than our method. 
As pre-trained sequence-to-sequence language models are good at copying input and generating natural-sounding output, we hypothesize that our approach will better allow us to change connotative meaning without affecting fluency and denotation. In contrast, approaches such as ``vocab boosting'' \cite{ghosh-etal-2017-affect} increase the logits of key connotative words, which would necessarily decrease the probabilities of functional words and words necessary for maintaining denotative meaning. Other approaches such as reinforcement learning \cite{pasunuru2017reinforced} may further decrease these desired qualities, while trying to maximize another objective.

\section{Evaluation Tasks and Test Data}
To evaluate our methods for argument reframing we need to look beyond our automatically labeled data. We consider two tasks: 1) reframing an argument that contains partisan language to a less partisan argument; and 2) reframing an appeal to fear or prejudice fallacy to an argument without this fallacy. 

Recently \citet{webson-etal-2020-undocumented}, proposed resources and methods to disentangle denotation and connotation in vector spaces. They evaluate their methods on a sample of around 300 collocations from vocabulary of Congressional records \cite{gentzkow2019measuring} and Hyper-partisan News \cite{kiesel-etal-2019-semeval} that occur at least 100 times and have high partisan skew. We use these words to filter arguments from the subreddits \textit{ChangeMyView} and \textit{Politics}. Some of these collocations include phrases such as \textit{abortion providers, investment vehicles, broken system, soft power,} and \textit{territorial claims}.
We randomly sample 100 such arguments to benchmark the performance of our model and further use towards human evaluation. 

In addition, we test our models on propaganda techniques employed in news articles with an \textit{Appeal to Fear or Prejudice} \cite{da2019fine}. There are a total of 182 sentence-level text fragments labeled as Appeal to Fear or Prejudice in the dataset released by \citet{da2019fine}. We classify these 182 fragments as claims/premises/non-argument and randomly sample 50 premises. Our goal is to reduce the fallacious nature of the argument without changing the denotative meaning. 

As our training distribution is different from these two datasets, these tasks and test sets allow us to better test the generalization capabilities of our models. Furthermore, almost none of the collocations introduced by \citet{webson-etal-2020-undocumented} appear in the connotation dictionary of \citet{allaway2020unified}, which helps us avoid the risk of mimicking replacements from our training data.

\begin{table}[]
\small
\centering
\begin{tabular}{|l|l|}
\hline
INP1 &  \begin{tabular}[c]{@{}l@{}}It would be \textit{\color{red}dangerous , suicidal folly} for \\infidels to pretend that ramadan is not the \\month of jihad\end{tabular} \\ \hline
HUM1 & \begin{tabular}[c]{@{}l@{}}It would be \textit{\color{blue}counterproductive and}\\ \textit{\color{blue}unreasonable} for infidels to ........ jihad\end{tabular} \\ \hline
INP2 & \begin{tabular}[c]{@{}l@{}}Trump backs away from further \\ \textit{\color{red}military confrontation} with Iran\end{tabular}     \\ \hline
HUM2 & \begin{tabular}[c]{@{}l@{}}Trump backs away from further \\ \textit{\color{blue}military} \textit{\color{blue}engagement} with Iran\end{tabular}        \\ \hline
\end{tabular}
\caption{INP1 and INP2 are test data instances where INP1 is a Appeal to Fear example while INP2 is an argument containing partisan collocation }
\label{table:test}
\end{table}

For both of these tasks, we ask humans to generate reframings based on our input test data for comparison and benchmark. We recruit two experts with argumentation and journalism background (not authors of this paper) to reframe arguments. For Appeal to Fear the instructions given were to make it less fallacious by reducing the fear and rephrasing the argument (HUM1 in Table \ref{table:test}), while for arguments with partisan collocation the human was instructed to change the collocation so as to make it trustworthy (HUM2 in Table \ref{table:test}).

\section{Experimental Setup} \label{section:exp}
To compare the quality of the reframed arguments, we benchmark our ENTRUST model  against human performance and four baseline systems described below. For the data containing collocations from \cite{webson-etal-2020-undocumented}, because we know they represent partisan language the ideal goal is to reframe them. For Appeal to Fear or Prejudice data we reframe words which portray an emotion of fear based on the popular NRC Emotion Lexicon \cite{mohammad2013crowdsourcing}.

\subsection{Baseline Systems} 
As argument reframing is a new task, we adapt several baselines that  have been used for other generation tasks and also compare with human-generated reframings.

\paragraph{\textbf{Bart wIthout Demarcator and ENtailment (BART$^{w/o}_{D+EN}$)}:}
This is the pre-trained BART model fine-tuned on our parallel data without explicitly adding signals on what to edit or reframe and without post-processing based on entailment scores. This experiment helps us understand if BART learns to adapt to the emotional connotations and can automatically edit partisan collocations or words inducing fear without control.
\paragraph{\textbf{Bart without EnTAilment (BART$^{w/o}_{EN}$)}:}
This is the pre-trained BART model fine-tuned on our parallel data with explicit signals ([SEP] token) but without the NLI component as a post-processing tool. This experiment helps us understand how well BART learns to adapt to the emotional connotations without altering the denotative meaning once guided with what to reframe.
\paragraph{\textbf{Lexical Replacement (LEXREP)}:} We use a similar method employed for our parallel data creation. We rely on Masked Language Models for lexical substitutions. Because our goal is to reframe arguments to be trustworthy we prefer substitutions which have a connotation of trust in the resource by \citet{allaway2020unified}. In case we cannot find the substitution in the connotation dictionary we honor default MLM predicted infilling.

\paragraph{\textbf{Generative Style Transformer (GST)}:} We use the state of art for text style transfer by \citet{sudhakar2019transforming}, which is a part of a larger ``\textit{Delete Retrieve Generate}" framework \cite{li2018delete}. To maintain parity with other baselines, instead of letting the model delete attribute keywords we delete the partisan collocations or fear related words from the arguments as the first step, followed by the usual retrieve and generate steps. Our training data for this method includes only arguments labeled with their attribute (e.g., positive or negative). Arguments containing lexical connotations catering to trust are positive, while those not catering to trust are negative.

\begin{table}[]
\small
\centering
\begin{tabular}{|l|l|l|}
\hline
System  & Partisan Task & Appeal to Fear Task \\ \hline
BART$^{w/o}_{D+EN}$   & 64.1     & 38.5           \\ \hline
BART$^{w/o}_{EN}$   & 91.9     & 43.1           \\ \hline
GST     & 86.4     & 38.3           \\ \hline
LEXREP  & 92.4     & \underline{44.3}          \\ \hline
ENTRUST & \underline{92.9}     & \textbf{44.5}           \\ \hline\hline
HUMAN   & \textbf{93.9}     & 41.6 *          \\ \hline
\end{tabular}
\caption{Semantic Similarity of reframed arguments with input arguments. (*) Here human did not restrict themselves to just lexical framing, so automated metrics might penalize them for more reframing.}
\label{table:simscore}
\end{table}

\begin{table}[]
\small
\centering
\begin{tabular}{|p{1.5cm}|p{1.1cm}|p{1cm}|l|l|}
\hline
System  & Fluency & Meaning  & \vline Trust $\uparrow$ & Fear$\downarrow$ \\ \hline
INPUT   & -    & -   & \vline 3.24  &   3.36   \\ \hline
BART$^{w/o}_{D+EN}$   & 2.78    & 2.56   & \vline2.60 & 3.01   \\ \hline
BART$^{w/o}_{EN}$   & 3.39    &  3.00  & \vline 3.13 &  2.58   \\ \hline
LEXREP  & 3.38    &  3.00  & \vline3.08 &    2.54  \\ \hline
GST     & 2.14    &  1.81   & \vline 2.01 &  \underline{2.44}   \\ \hline
ENTRUST & \underline{3.51 }   & \underline{3.30}  & \vline \underline{3.52} &    \textbf{2.39}   \\ \hline\hline
HUMAN   & \textbf{3.72}    & \textbf{3.63}  & \vline \textbf{3.71} &  2.59     \\ \hline
\end{tabular}
\caption{Fluency and Meaning Preservation scores given by human judges on a scale of (1-5) for reframed arguments with respect to input arguments. Fluency and Meaning Preservation ratings are for all arguments in test set, while Trust ratings are for arguments with Partisan collocation (higher scores better), and Fear ratings for Appeal To Fear or Prejudice ones only (lower scores better).}
\label{table:human}
\end{table}

\subsection{Evaluation Criteria}

\paragraph{Automatic evaluation.}
One important criterion is to measure if the reframed arguments are faithful to the input. Even though we are changing the argument for connotations, it should still maintain the same denotative meaning as the input. To this end we calculate \textbf{Semantic Similarity} with our input using SENTENCE BERT(SBERT) \cite{reimers2019sentence}.  

\paragraph{Human evaluation.}
We use Amazon Mechanical Turk to evaluate on a total of 900 utterances, 750 generated from 5 systems and 150 utterances generated by humans.  We proposed a set of 3 criteria to evaluate the generated output: (1) \textbf{Fluency (F)} (``How fluent and grammatical are the utterances?''), 
(2) \textbf{Meaning Preservation (M)} (``How well does the reframed argument capture the same denotative meaning as the input argument?''), (3) \textbf{Trustworthiness/Presence of Fear(T/PF)}. For the 100 input arguments reflecting partisan view we ask Turkers to rate reframed arguments based on trustworthiness with respect to the input. For the 50 Appeal to Fear or Prejudice fallacies we ask Turkers to rate reframed arguments based on presence of fear (the intention behind this being that we want to rank systems which portray the least amount of fear). In both of these ratings we still ask Turkers to keep into account the denotative meaning (i.e., making it trustworthy or less fallacious at the expense of meaning alterations should be scored lower). We hired 40, 25, 39 (23 and 16) Turkers for the three separate tasks respectively. The computed IAA using Krippendorff's alpha for Fluency, Meaning Preservation , Trust-Worthiness and Presence of Fear is 0.62, 0.65, 0.51, 0.46, respectively.

\begin{table*}[!ht]
\small
\centering
\begin{tabular}{|@{ }p{3.8cm}@{ }|l@{ }|@{ }p{7.8cm}@{ }|@{ }l@{ }|@{ }l@{ }|@{ }l@{ }|}
\hline
Original Argument & System & reframed Argument & F & M & T$\uparrow$/PF$\downarrow$   \\ \hline
\multirow{6}{*}{\begin{tabular}[c]{@{}l@{}} It is difficult to think of any\\ single act that would do \\more to restore America's\\ \textit{\color{red}soft power} than the election\\ of Obama to the presidency \end{tabular}} & BART$^{w/o}_{D+EN}$  & \begin{tabular}[c]{@{}l@{}}  It is difficult to think of any single act that would do more to \\restore America's \textit{\color{red}soft power} than the election of Obama to \\the presidency   \end{tabular}  & 3.7 & 2.3  & 3.3  \\ 
    \cline{2-6}
    & BART$^{w/o}_{EN}$  & \begin{tabular}[c]{@{}l@{}}  It is difficult to think of any single act that would do more to \\restore America's \textit{\color{blue}moral authority} than the election of Obama\\ to the presidency  \end{tabular}  &3.7 & 3.3 & 3.0\\
    \cline{2-6}
    & LEXREP & \begin{tabular}[c]{@{}l@{}}  It is difficult to think of any single act that would do more to \\restore America's \textit{\color{blue}moral standing} than the election of Obama\\ to the presidency \end{tabular}  & 3.7  & 3.3  & 2.7 \\
    \cline{2-6}
    & GST & \begin{tabular}[c]{@{}l@{}} \textit{\color{blue}Hated} it is difficult to think of any single act that would do\\ more to restore America's \textit{\color{blue}economy} than the election of\\ Obama to the presidency \end{tabular}  & 1.7 & 1.7 & 2.7  \\
    \cline{2-6}
    & ENTRUST & \begin{tabular}[c]{@{}l@{}}  It is difficult to think of any single act that would do more to \\restore America's \textit{\color{blue}diplomatic credibility} than the \\election of Obama to the presidency \end{tabular}  & \textbf{4.3} & \textbf{3.7}  & \textbf{5.0} \\
    \cline{2-6}
    & HUMAN &  \begin{tabular}[c]{@{}l@{}}  It is difficult to think of any single act that would do more to \\restore America's \textit{\color{blue}political and economic influence} than the \\election of Obama to the presidency \end{tabular}  & 4.0 & 3.3 & 4.0 \textbf{}  \\
    \cline{2-6}
    \hline\hline
    
\multirow{6}{*}{\begin{tabular}[c]{@{}l@{}} Or he can \textit{\color{red}seize} this \\opportunity and use his\\ \textit{\color{red}government's} diplomatic\\ \textit{\color{red}influence} to \textit{\color{red}defend} the life of\\ an australian citizen , whose\\ \textit{\color{red}courageous} public service is \\recognised by countless \\people across the world \end{tabular}} & BART$^{w/o}_{D+EN}$  & \begin{tabular}[c]{@{}l@{}}  Or he can \textit{\color{red}seize} this opportunity and use his \textit{\color{red}governments's} \\diplomatic \textit{\color{red}influence} to \textit{\color{red}defend} the life of an australian citizen \\, whose courageous \textit{\color{blue} military} service is recognised by\\ countless people across the world   \end{tabular}  & 2.7 & 3.0  & 2.3  \\ 
    \cline{2-6}
    & BART$^{w/o}_{EN}$  & \begin{tabular}[c]{@{}l@{}}  Or he can \textit{\color{blue}honor} this opportunity and use his \textit{\color{blue}nation's} \\diplomatic \textit{\color{blue}resources} to \textit{\color{blue}honor} the life of an australian citizen \\, whose  \textit{\color{blue} unwavering} public service is recognised by countless \\people across the world   \end{tabular} & 4.0  &  4.7 &  \textbf{1.3}\\
    \cline{2-6}
    & LEXREP & \begin{tabular}[c]{@{}l@{}}  Or he can \textit{\color{blue}take} this opportunity and use his \textit{\color{blue}nations's} \\diplomatic \textit{\color{blue}resources} to \textit{\color{red}defend} the life of an australian citizen \\, whose courageous \textit{\color{blue} military} service is recognised by \\countless people across the world   \end{tabular}  & \textbf{4.7}  & 4.0  & 2.0\\
    \cline{2-6}
    & GST & \begin{tabular}[c]{@{}l@{}}  Or he can \textit{\color{blue}do} this opportunity and use his \\diplomatic \textit{\color{blue}expertise} to \textit{\color{red}change} the life of an australian citizen \\, whose public service is recognised by countless \\people across the world   \end{tabular}  &  1.7 & 2.0  &2.0  \\
    \cline{2-6}
    & ENTRUST & \begin{tabular}[c]{@{}l@{}}  Or he can \textit{\color{blue}honor} this opportunity and use his \textit{\color{blue}nation's} \\diplomatic \textit{\color{blue}resources} to \textit{\color{blue}vindicate} the life of an australian\\ citizen , whose \textit{\color{blue}unwavering} public service is recognised \\by countless people across the world   \end{tabular} & \textbf{4.7}  & \textbf{4.7}  & \textbf{1.3}   \\
    \cline{2-6}
    & HUMAN & \begin{tabular}[c]{@{}l@{}}Or he can \textit{\color{blue}pick up} this opportunity and use his government’s\\diplomatic influence to defend the life of an Australian\\ citizen, whose \textit{\color{blue}actions have been publicly recognized as} \\\textit{\color{blue}highly relevant at an international level}. 
\end{tabular}  & 3.7  & 4.0  &   2.3  \\
    \cline{2-6}
    \hline
\end{tabular}
\caption{Examples of generated outputs from different systems (with human reframed argument as references) for arguments containing partisan collocations and appeal to fear, respectively. We show average scores (over three annotators) on a 1-5 scale with 1 denotes the worst and 5 be the best. More examples in Appendix}
\label{table:all}
\end{table*}

\section{Results}
\paragraph{Automatic Evaluation.}
As can be seen in Table \ref{table:simscore} our model ENTRUST maintains the denotative meaning with the input better than other systems ($p<.001$ using approximate randomization tests) and only marginally behind humans when it comes to arguments with partisan collocations. For Appeal to Fear or Prejudice our system maintains better denotative meaning than all systems except LEXREP ($p<.001$). The automatic metric somewhat penalizes humans for changing more content than just targeted words; this unreliability is a known issue with  automated metrics \cite{novikova-etal-2017-need} and strongly implies a need for human evaluation.

\paragraph{Human Evaluation.}
Table \ref{table:human} shows the results of our human-based evaluations. For fluency, meaning preservation, trustworthiness, and reduction of fear the ENTRUST model is better than all the baselines ($p<.001$ using approximate randomization tests). It is further encouraging to see that the entailment step helps us maintain better denotative meaning (See Table \ref{table:human} Col3: Row 4 vs Row 7). For Presence of Fear, Turkers often rate our ENTRUST model to be the least fearful, including slightly when compared to reframings of an expert. We hypothesize this is because the human judges found it difficult to completely remove fear while keeping the denotative meaning (indeed the humans scores slightly better on meaning comparing with our system ENTRUST). 
Sometimes, an ungrammatical generation or a reframing which change the meaning will contain less fear (rating 1 meaning no fear at all). However, to avoid this we explicitly asked Turkers to rate those samples as moderate so as to not bias the overall results.

\section{Analysis and Discussion}

As can be seen in Table \ref{table:all}, the ENTRUST model accurately captures \textit{diplomatic credibility} as an alternate for \text{soft power} which is encouraging as soft power is measured through \textit{culture, diplomacy, education, business/innovation, and government.}\footnote{\url{https://en.wikipedia.org/wiki/Soft_power}} The BART$^{w/o}_{D+EN}$ model often fails to reframe anything, which shows the importance of adding [SEP] tokens as explicit supervision so that the model knows what to edit. The GST model fails at both grammaticality and meaning preservation,  which makes it harder to judge its trustworthiness and ability to ameliorate fearful appeal. Finally, ENTRUST reframings are not static. Table \ref{table:context} shows that for the same collocation of \textit{targeted killing}, the reframings are different, contingent on the context. This goes on to prove that our model not only generalizes to unseen test data, but can produce novel, grammatical and meaningful edits based on context.

\begin{table}[]
\small
\centering
\begin{tabular}{|l|}
\hline
\begin{tabular}[c]{@{}l@{}}An Iranian government official seemed to suggest that \\President Trump’s properties  could be potential targets\\ in retaliation for the US \textit{\color{red}targeted killing} of Iranian\\ general Qassem Soleimani.\end{tabular}                  \\ \hline
\begin{tabular}[c]{@{}l@{}}An Iranian government official seemed to .... suggest that \\President Trump’s properties .... in retaliation for the US\\ \textit{\color{blue}involvement in the execution} of Iranian .....\end{tabular}     \\ \hline\hline
\begin{tabular}[c]{@{}l@{}}A federal appeals court ordered the U.S. Department of \\Justice to turn over key portions of a memorandum\\ justifying the government's \textit{\color{red}targeted killing} of people \\linked to terrorism, including Americans\end{tabular}        \\ \hline
\begin{tabular}[c]{@{}l@{}}A federal appeals court ordered the U.S. Department of \\Justice to turn ......... justifying the government's \\\textit{\color{blue}extrajudicial execution} of people ... terrorism, .........\end{tabular} \\ \hline
\end{tabular}
\caption{Different reframed arguments by ENTRUST model based on same collocation}
\label{table:context}
\end{table}

\section{Related Work} \label{sec:related}

The effects of lexical framing have been studied for social and political issues, although our work is the first to use lexical framing in generation for positive framing effects (less partisan, no appeal to fear fallacy). \citet{demszky2019analyzing} and \citet{tyagi2020computational} study political polarization and how this manifests in differences in word choice among different groups; \citet{khudabukhsh2020we} provide an interpretable framework using machine translation between groups to generate differences.
While these works encourage computational approaches  to reframe arguments for better lexical choice, these approaches do not control for denotation or connotation and thus may cause differences in word choice to result in a change in meaning.
The most similar work to ours is that of \citet{pryzant2020automatically}, who use a corpus of Wikipedia edits to train a model for debiasing, which includes framing.
However, in their work communicative intent is left implicit; the corpus is only labeled for types of debiasing, which includes framing at a high level and not the connotations involved. Thus, their model only learns lexical differences, whereas our model is controllable.  

While our focus is on lexical framing, other work has investigated the identification of other types of frames and their effects. \newcite{greene-resnik-2009-words} studied \textit{syntactic} framing, finding a link between implicit sentiment and syntactic packagings. 
Previous studies have also involved \textit{emphasis} framing -- \newcite{ding-pan-2016-personalized} find that emphasizing aspects of products given personal information is more effective for content selection in advertisements. Other research has involved \textit{issue} framing -- \newcite{ajjour-etal-2019-modeling} and \newcite{hartmann-etal-2019-issue} study how arguments are framed in debates (e.g., in terms of economics or safety). \citet{Nguyen2013LexicalAH} and \citet{field2018framing} study ``agenda-setting'' for news and congressional debates and \newcite{DBLP:journals/pacmhci/AugustOTSR18} for study recruitment. \newcite{cano-basave-he-2016-study} and \newcite{Musi2019FramingFS} leverage \textit{semantic} frames for distant labeling and analysis of arguments in political debates, respectively, and find, for example, that evidence and reasoning are among the most common. However, these approaches have focused on identification rather than generation. 

Finally, our work is also related to style transfer and controllable generation. Much of the work in ``style transfer'' has referred to changing the sentiment of a statement, which changes the truth condition and thus the denotative meaning. Sentiment is often explicitly marked and thus approaches such as deleting and replacing lexical markers are effective \cite{li-etal-2018-delete, sudhakar-etal-2019-transforming}, although our experiments showed the difficulty of applying these techniques to our task.
To control text generation by limiting contradictions, \citet{pasunuru2017reinforced} use an entailment score as a reward in Reinforcement Learning, ensuring that a generated text is logically implied 
 by the ground-truth text. \citet{holtzman2018learning} utilize a discriminative model trained on SNLI \cite{bowman2015large} to complement an RNN generator and guide the decoding process to improve contradictions in generation. Although we experimented with both of these approaches, including the approach of \citet{holtzman2018learning}  with MNLI to account for entailment in text generation, none of them yielded better results than our method. Other approaches have explored ``vocab boosting'' \cite{ghosh-etal-2017-affect} for tasks such as de-biasing \cite{ma2020powertransformer}, which involves increasing the values of certain words; however, as these values are on the simplex, the softmax function necessarily decreases the values of other logits which are key to fluency such as function words.

\section{Conclusion}
Our experiments showed that our approach is effective in reframing partisan arguments and appeals to fear for increased trustworthiness. We provided a method for creating a dataset using a lexical resource for connotations and masked language modeling. We used this dataset to fine-tune  a controllable text generation model for the task of changing connotative meaning and used a model trained for natural language inference to maintain the denotative meaning. Our evaluations found that our approach generalized to two different tasks and data sets.
In future work, we plan to directly incorporate the role of stance in framing (for arguments and counter-arguments). We also plan to expand 
our work  to generating concessions \cite{Musi2018HowDY}, where the goal is for the speaker to portray some point of agreement in a positive light before disagreeing.

\section{Ethics}
Our data is collected from Reddit and we understand and respect user privacy. Our models are fine-tuned on sentence level data obtained from user posts. These do not contain any explicit detail which leaks information about a users name, health, negative financial status, racial or ethnic origin, religious or philosophical affiliation or beliefs, sexual orientation, trade union membership, alleged or actual commission of crime. 

Second, although we use language models trained on data collected from the Web, which have been shown to have issues with bias and abusive language \cite{sheng-etal-2019-woman, wallace-etal-2019-universal}, the inductive bias of our models should limit inadvertent negative impacts. Unlike model variants such as GPT, BART is a conditional language model, which provides more control of the generated output. We have two levels of control on our generation approach: lexical replacements via connotations associated with trust and an entailment method that aims to keep the same denotation of the original argument.  
While dual-use concerns are certainly possible here, we think that open-sourcing this technology will help to generate arguments with more balanced and trusted language that are less targeted towards partisanship or appeals to fear. 

Finally, while there may be concerns about building generative models for persuasion, social scientists distinguish persuasion from manipulation based on two aspects: dissimulation and constraint \cite{Nettel2012}.  Dissimulation involves concealing intention, which requires hiding information, whereas constraint involves removing options from the audience and forcing them to accept the conclusion.
Our work on reframing arguments does not aim to hide information about a topic or present it as the only choice, but aims to provide the same argument using more balanced and trusted language. We achieve this by two key components of our technology: controllable text generation (connotation associated with trust) and entailment model to ensure same denotation. 

The technology should be used responsibly, particularly making sure the generation is controllable for trust and positive emotion and that the entailment component is used for ensuring the same denotation with the original argument. 

Finally we pay the Turkers at a rate of 15\$/hour, complying with minimum wage standards in most places.

\bibliography{anthology,custom}
\bibliographystyle{acl_natbib}

\appendix
\section{Hyper-Parameters and Other Experimental Settings}

\begin{enumerate}
    \item{\textbf{No of Parameters:}} For Connotation guided MLM and Entailment we use RoBERTa large model (355M). For generation we use the BART large checkpoint (400M parameters) and use the implementation by FAIRSEQ \cite{ott2019fairseq} 
     \footnote{\url{https://github.com/pytorch/fairseq/tree/master/examples/bart}}. 
    \item{\textbf{No of Epochs:}}  We fine-tune pre-trained BART for 20 epochs for ENTRUST model and save best model based on validation perplexity.
    \item{\textbf{Training Time:}} Our training time is 80 minutes for BART.
    \item{\textbf{Hardware Configuration:}} We use 4 RTX 2080 GPU
    \item{\textbf{Training Hyper parameters:}} We use the same parameters mentioned in the github repo where BART was fine-tuned for CNN-DM summarization task with the exception of MAX-TOKENS (size of each mini-batch, in terms of the number of tokens.) being 1024 for us. For entailment we rely on AllenNLP roberta checkpoint that is finetuned on MNLI \footnote{url{https://demo.allennlp.org/textual-entailment}}
    \item{\textbf{Decoding Strategy \& Hyper Parameters:}}For decoding we reframe from our models using a top-k random sampling scheme \cite{fan2018hierarchical}. At each timestep, the model generates the probability of each word in the vocabulary being the likely next word. We randomly sample from the k = 5 to k=50 for most likely candidates from this distribution and re-rank them based on entailment scores.
\end{enumerate}

\subsection{Role of Context in re-framing}
As mentioned ENTRUST reframings are not static. Table \ref{table:context} shows that for multiple same collocations the reframings are different, contingent on the context. This supports our claim that our model not only generalizes on unseen test data, but can produce novel, grammatical and meaningful edits based on context. We notice that for the first example, ENTRUST believes scientific proof leads to be better credibility instead of using the broad umbrella term experts. For the same collocation of \textit{leading expert} we see that given the premise talks about a neurological basis related to birth , ENTRUST reframes it to reflect to \textit{leading medical fact} as the term ``expert" can be ambiguous, while people are generally more likely to believe medical facts.
\begin{table}[]
\small
\centering
\begin{tabular}{|l|}
\hline
\begin{tabular}[c]{@{}l@{}}Statistically based-wise though, the number of violent \\crime is more prevalent in warm areas,\textit{\color{red}leading experts} \\to believe a correlation between warm and people being \\more aggressive.\end{tabular}                  \\ \hline
\begin{tabular}[c]{@{}l@{}}Statistically based-wise though, the number of violent \\crime is more prevalent in warm areas,\textit{\color{blue}leading scientific} \\ \textit{\color{blue}proof} to believe a correlation between warm and people\\ being more aggressive.\end{tabular}     \\ \hline\hline
\begin{tabular}[c]{@{}l@{}}The \textit{\color{red}leading experts} all agree that there's a neurological \\basis \& there is significant evidence showing it's caused\\ during fetal neurological development\end{tabular}        \\ \hline
\begin{tabular}[c]{@{}l@{}}The \textit{\color{blue}leading medical facts} all agree that there's a \\neurological basis \& there is significant evidence showing\\ it's caused during fetal neurological development\end{tabular}  \\ \hline
\begin{tabular}[c]{@{}l@{}}We need leaders who recognize our \textit{\color{red}moral obligation} \\ for health equality\end{tabular}        \\ \hline
\begin{tabular}[c]{@{}l@{}}We need leaders who recognize our \textit{\color{blue}moral accountability} \\ for health equality\end{tabular}        \\ \hline
\begin{tabular}[c]{@{}l@{}}We all have a \textit{\color{red}moral obligation} to the next  generation to \\leave America’s natural resources in better condition than\\ when we inherited them\end{tabular}        \\ \hline
\begin{tabular}[c]{@{}l@{}}We all have a \textit{\color{blue}personal oath} to the next  generation to \\leave America’s natural resources in better condition than\\ when we inherited them\end{tabular}        \\ \hline

\end{tabular}
\caption{Different reframed arguments by ENTRUST model based on same collocation}
\label{table:context}
\end{table}

\begin{table*}[!ht]
\small
\centering
\begin{tabular}{|@{ }p{3.8cm}@{ }|l@{ }|@{ }p{7.8cm}@{ }|@{ }l@{ }|@{ }l@{ }|@{ }l@{ }|}
\hline
Original Argument & System & reframed Argument & F & M & T$\uparrow$/PF$\downarrow$   \\ \hline
\multirow{6}{*}{\begin{tabular}[c]{@{}l@{}} AIPAC called on the Obama \\administration to take steps to\\ defuse tension with Israel,\\ while Israel's ambassador to \\the U.S. said \textit{\color{red}bilateral} \\\textit{\color{red}relations} are in crisis \end{tabular}} & BART$^{w/o}_{D+EN}$  & \begin{tabular}[c]{@{}l@{}}  AIPAC called on the Obama administration to take steps to \\defuse tension with Israel, while Israel's ambassador to the \\U.S. said \textit{\color{red}bilateral relations} are in \textit{\color{blue}fact stable} \end{tabular}  & 3.7 & 2.0  & 2.3  \\ 
    \cline{2-6}
    & BART$^{w/o}_{EN}$  & \begin{tabular}[c]{@{}l@{}}  AIPAC called on the Obama administration to take steps to \\defuse tension with Israel, while Israel's ambassador to the \\U.S. said \textit{\color{blue}diplomatic relations} are in crisis \end{tabular}  &4.7 & 3.3 & 4.3\\
    \cline{2-6}
    & LEXREP &\begin{tabular}[c]{@{}l@{}}  AIPAC called on the Obama administration to take steps to \\defuse tension with Israel, while Israel's ambassador to the \\U.S. said \textit{\color{blue}that negotiations} are in crisis \end{tabular}  & 4.0  & 2.3  & 3.0 \\
    \cline{2-6}
    & GST & \begin{tabular}[c]{@{}l@{}}  AIPAC called on the Obama administration to take steps to \\defuse tension with Israel, while Israel's ambassador to the \\ \textit{\color{blue}law said} are in crisis \end{tabular}  & 1.3 & 1.3 & 1.3  \\
    \cline{2-6}
    & ENTRUST & \begin{tabular}[c]{@{}l@{}}  AIPAC called on the Obama administration to take steps to \\defuse tension with Israel, while Israel's ambassador to the \\U.S. said \textit{\color{blue}diplomatic ties} are in crisis \end{tabular}  & \textbf{5.0} & \textbf{4.7}  & \textbf{4.7} \\
    \cline{2-6}
    & HUMAN &  \begin{tabular}[c]{@{}l@{}}  AIPAC called on the Obama administration to take steps to \\defuse tension with Israel, while Israel's ambassador to the \\U.S. said \textit{\color{blue}relations} are in crisis \end{tabular}  & 4.7 & 3.7 & 4.0 \textbf{}  \\
    \cline{2-6}
    \hline\hline
    
\multirow{6}{*}{\begin{tabular}[c]{@{}l@{}} Kentucky Senate President\\ Robert Stivers on Wednesday\\ a federal judge to withhold \\action against Davis for her \\defiance of the ruling by \\denying \textit{\color{red}marriage licenses }to \\gay couples, until the \\Kentucky General Assembly \\can act \end{tabular}} & BART$^{w/o}_{D+EN}$  & \begin{tabular}[c]{@{}l@{}}  Kentucky Senate President Robert Stivers on Wednesday \\asked a federal judge to withhold action against Davis for her\\ defiance of the \textit{\color{blue}law} ruling by denying \textit{\color{red} marriage licenses}\\ to gay couples, until the Kentucky General Assembly can act.
   \end{tabular}  & 4.3 & 3.7  & 3.0 \\ 
    \cline{2-6}
    & BART$^{w/o}_{EN}$  & \begin{tabular}[c]{@{}l@{}}  Kentucky Senate President Robert Stivers on Wednesday \\asked a federal judge to withhold action against Davis for her\\ defiance of the ruling by denying \textit{\color{blue}legal marriage equality}\\ to gay couples, until the Kentucky General Assembly can act.
   \end{tabular} & \textbf{4.7}  &  4.0 &  4.0 \\
    \cline{2-6}
    & LEXREP & \begin{tabular}[c]{@{}l@{}}  Kentucky Senate President Robert Stivers on Wednesday \\asked a federal judge to withhold action against Davis for her\\ defiance of the ruling by denying \textit{\color{blue}wedding services}\\ to gay couples, until the Kentucky General Assembly can act.
   \end{tabular}  & \textbf{4.0}  & 1.3  & 2.3\\
    \cline{2-6}
    & GST & \begin{tabular}[c]{@{}l@{}}  Kentucky Senate President Robert Stivers on Wednesday \\asked a federal judge to withhold action against Davis for her\\ defiance of the ruling by denying \textit{\color{blue}the}\\ gay couples, until the Kentucky General Assembly can act.
   \end{tabular} &  1.7 & 1.3  &2.7  \\
    \cline{2-6}
    & ENTRUST & \begin{tabular}[c]{@{}l@{}}  Kentucky Senate President Robert Stivers on Wednesday \\asked a federal judge to withhold action against Davis for her\\ defiance of the ruling by denying \textit{\color{blue}legal marriage authorization}\\ to gay couples, until the Kentucky General Assembly can act.
   \end{tabular} & 4.3  & \textbf{4.7}  & \textbf{4.3}   \\
    \cline{2-6}
    & HUMAN & \begin{tabular}[c]{@{}l@{}}  Kentucky Senate President Robert Stivers on Wednesday \\asked a federal judge to withhold action against Davis for her\\ defiance of the ruling by denying \textit{\color{blue} marriage paperwork}\\ to gay couples, until the Kentucky General Assembly can act.
   \end{tabular}  & \textbf{4.7}  & 3.0  &   3.7  \\
    \cline{2-6}
    \hline\hline
    \multirow{6}{*}{\begin{tabular}[c]{@{}l@{}} A campaign of \text{\color{red} hate} laced with\\ \text{\color{red} blatant} anti-semitic overtones \end{tabular}} & BART$^{w/o}_{D+EN}$  & \begin{tabular}[c]{@{}l@{}}  A campaign of \text{\color{red} hate} laced with \text{\color{blue} genuine} anti-semitic overtones \end{tabular} & \textbf{4.3}  &  \textbf{4.7}  & 2.3  \\ 
    \cline{2-6}
    & BART$^{w/o}_{EN}$  & \begin{tabular}[c]{@{}l@{}}  A campaign of \text{\color{blue} trust} laced with \text{\color{blue} obvious} anti-semitic overtones \end{tabular}  & 2.0  & 1.7  &  \textbf{2.0}\\
    \cline{2-6}
    & LEXREP &\begin{tabular}[c]{@{}l@{}}  A campaign of \text{\color{blue}intimidation} laced with \text{\color{blue}strong} anti-semitic \\overtones \end{tabular}  & 3.3  &  3.0  &  3.0 \\
    \cline{2-6}
    & GST & \begin{tabular}[c]{@{}l@{}}  A campaign of \textit{\color{blue}injuring rollercoaster} laced with anti - semitic \\overtones \end{tabular} & 2.3  & 1.0  &  3.0  \\
    \cline{2-6}
    & ENTRUST & \begin{tabular}[c]{@{}l@{}}   A campaign of \textit{\color{blue}prejudice } laced with \textit{\color{blue}genuine} anti - semitic\\ overtones \end{tabular}  & 3.7  & 4.0  & \textbf{2.0} \\
    \cline{2-6}
    & HUMAN &  \begin{tabular}[c]{@{}l@{}}A campaign with \textit{\color{blue}clear} anti-semitic overtones which\\ \textit{\color{blue}does not promote peace}\end{tabular}  & 2.7  & 2.7  &  3.7  \\
    \cline{2-6}
    \hline
\end{tabular}
\caption{Examples of generated outputs from different systems (with human reframed argument as references) for arguments containing partisan collocations and appeal to fear, respectively. We show average scores (over three annotators) on a 1-5 scale with 1 denotes the worst and 5 be the best.}
\label{table:all1}
\end{table*}


\end{document}